\documentclass{article}

\usepackage[preprint]{neurips_2024}

\usepackage[utf8]{inputenc} 
\usepackage[T1]{fontenc}    
\usepackage{hyperref}       
\usepackage{url}            
\usepackage{booktabs}       
\usepackage{amsfonts}       
\usepackage{nicefrac}       
\usepackage{microtype}      
\usepackage{xcolor}         
\usepackage{graphicx}
\usepackage{float}
\usepackage{amsmath} 
\usepackage{bbding}
\usepackage{authblk}
\setcitestyle{numbers,square}

\title{COMOGen: A Controllable Text-to-3D Multi-object Generation Framework}

\author[1]{\textbf{Shaorong Sun}}
\author[1]{\textbf{Shuchao Pang}}
\author[2]{\textbf{Yazhou Yao}}
\author[3]{\textbf{Xiaoshui Huang}}
\affil[1]{School of Cyber Science and Engineering, Nanjing University of Science and Technology}
\affil[2]{School of Computer Science and Engineering, Nanjing University of Science and Technology}
\affil[3]{Department of Computer Science and Engineering, Shanghai Jiao Tong University}

\begin{document}

\maketitle

\begin{abstract}
The controllability of 3D object generation methods is achieved through input text. Existing text-to-3D object generation methods primarily focus on generating a single object based on a single object description. However, these methods often face challenges in producing results that accurately correspond to our desired positions when the input text involves multiple objects.  
To address the issue of controllability in generating multiple objects, this paper introduces COMOGen, a \textbf{CO}ntrollable text-to-3D \textbf{M}ulti-\textbf{O}bject \textbf{Gen}eration framework. COMOGen enables the simultaneous generation of multiple 3D objects by the distillation of layout and multi-view prior knowledge. The framework consists of three modules: the layout control module, the multi-view consistency control module, and the 3D content enhancement module. 
Moreover, to integrate these three modules as an integral framework, we propose Layout Multi-view Score Distillation, which unifies two prior knowledge and further enhances the diversity and quality of generated 3D content. 
Comprehensive experiments demonstrate the effectiveness of our approach compared to the state-of-the-art methods, which represents a significant step forward in enabling more controlled and versatile text-based 3D content generation.
\end{abstract}

\section{Introduction}

With the continuous development of industries such as gaming, VR and AR, there is a growing demand for 3D content. However, manually creating 3D content is a very costly endeavor, making it urgent to solve the problem of quickly generating high-quality 3D content that meets people's needs using artificial intelligence technology. In the field of 2D image generation, diffusion model~\cite{ho2020denoising} have achieved astonishing success. Several methods \cite{stale-zero123,zheng2024point,huang2024frozen} capitalize on the existing success of 2D fields for 3D tasks. However, in the realm of 3D generation, achieving similar success to diffusion models is not straightforward. To solve this problem, DreamFusion~\cite{poole2022dreamfusion} proposes score distillation sampling, which distills pretrained prior knowledge from the 2D diffusion model into 3D implicit representations~\cite{barron2021mip}, but its generation of 3D content is slow, and it also faces the Janus problem. The representative of another approach is Instant3D~\cite{li2023instant3d}, with the emergence of the Objaverse~\cite{deitke2023objaverse,deitke2024objaverse}, it uses an improved transformer~\cite{zhang2022dino} architecture and a feed-forward training method to achieve end-to-end text-to-3D content generation, however, this approach requires substantial computational resources for model training in the 3D generation.

\begin{figure}[H]
\centering
\includegraphics[width=1.0\textwidth]{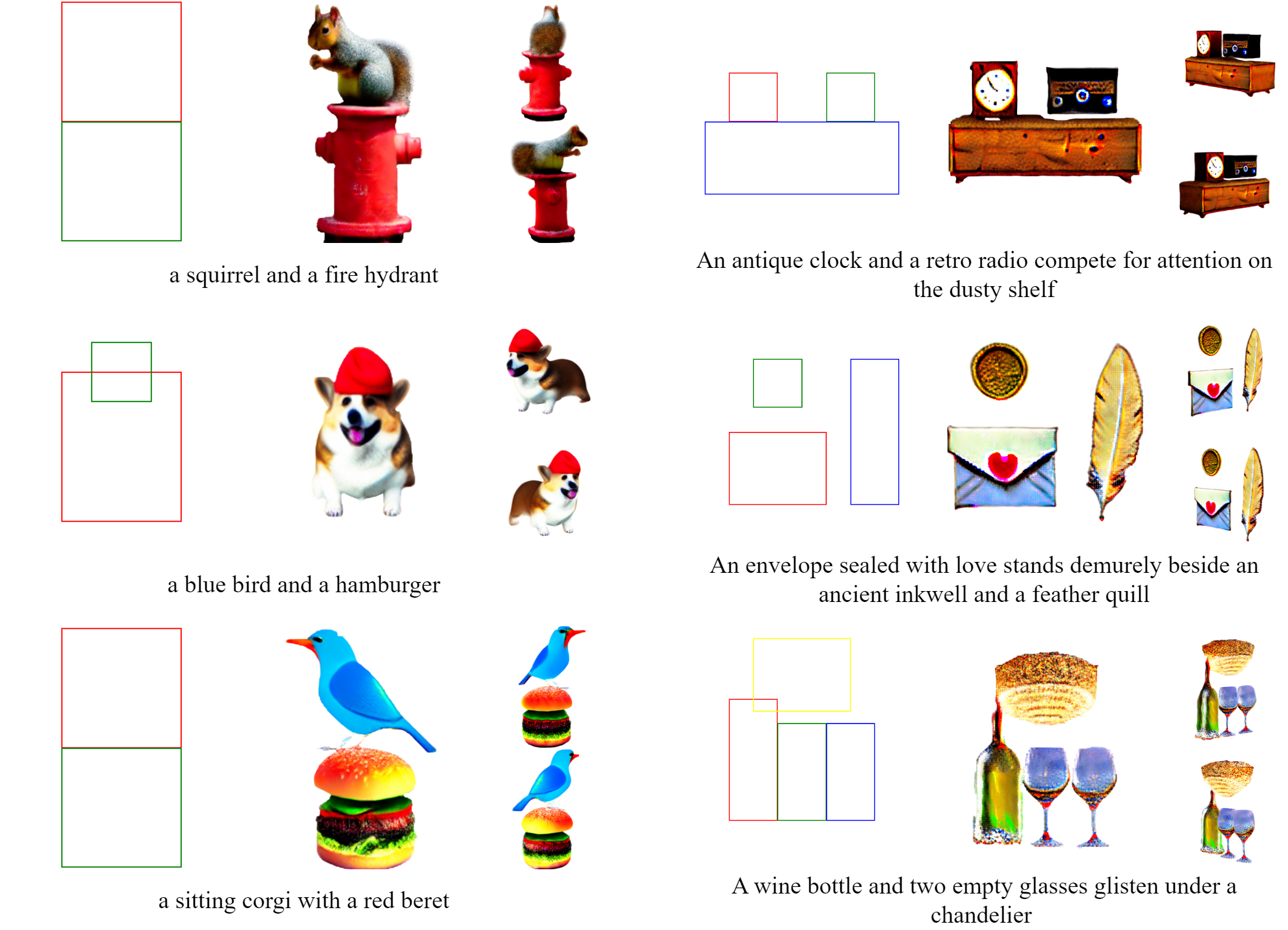}
\caption{Some samples generated by COMOGen from text and given any bounding boxes.}
\label{fig:1}
\end{figure}

More importantly, the existing works \cite{liu2023syncdreamer,zheng2024sketch3d,xu2023dmv3d,chen2023control3d,lin2023magic3d,liu2024uni3d} on text-to-3D object generation primarily focus on generating single objects, lacking control over the final outcome of the generated objects. For example, when a text prompt containing multiple objects is provided, like "three donuts," current methods for text-to-3D content generation struggle to produce satisfactory results. The positions of the generated 3D objects exhibit various possibilities, and the individual object positions fail to meet the desired requirements. Additionally, these methods often encounter challenges such as object fusion and omission. The final count of generated 3D objects is often incorrect, and the characteristics of multiple objects tend to get mixed together.

In this paper, we present a simple yet effective framework called COMOGen for generating multiple 3D contents while ensuring that the positions of the generated 3D objects align with our expectations using bounding boxes. The key idea is to leverage the strong prior knowledge from multiple diffusion models. We extend classical text-to-3D methods into three components: the layout control module, the multi-view consistency control module, and the 3D content enhancement module. Furthermore, we propose Layout Multi-view Score Distillation(LMSD) for unifying layout priors and multi-view priors. Specifically, as illustrated in Fig. \ref{fig:2}, for a given text prompt and bounding box, we first use a layout control module to generate diverse and reasonably arranged images as 2D priors for multi-view control modules. Then, in the layout control module, we propose Layout-SDS to distill layout pre-training knowledge guided by the input text and bounding boxes. In the multi-view consistency control module, we design the use of Multi-view SDS to distill multi-view consistency prior knowledge guided by 2D prior images. Furthermore, to unify the layout knowledge and multi-view knowledge obtained from the two diffusion models, we propose LMSD. Finally, we calculate the multiple loss values of Layout-SDS, Multi-view SDS, and LMSD to optimize the initial 3D content and obtain the final multiple 3D objects. Additionally, in the 3D content enhancement module, we fine-tune the diffusion models based on multi-view renderings of the optimizing 3D instances, which are used to fit the distribution represented by 3D content. 

COMOGen exhibits strong generalization capabilities, where the GLIGEN~\cite{li2023gligen} and Stable Zero123~\cite{stale-zero123} models used in our framework are trained based on stable-diffusion on the COCO~\cite{lin2014microsoft} dataset and Objaverse~\cite{deitke2023objaverse,deitke2024objaverse} dataset, respectively. Furthermore, COMOGen maintains creativity and diversity in inferring 3D information, allowing us to generate multiple layout-reasonable 3D objects from given text and layout. COMOGen enables anyone to generate reasonable 3D content simply from text and bounding boxes, potentially revolutionizing industries such as gaming, visual effects, film-making, and content creation.

Our main contributions are summarized as follows:
\begin{itemize}
\item We present a controllable text-to-3D generation framework capable of simultaneously generating multiple objects with reasonable spatial relationships. Our approach requires only text input and corresponding bounding boxes to generate 3D content, eliminating the need for any supervised text-to-3D data for training.
\item We propose Layout Multi-view Distillation Sample, a method used to unify layout priors and multi-view consistency priors distilled from pretrained diffusion models.
\item We have improved the quality of generation by using the iterative optimization framework COLA to fine-tune the diffusion model and fit the distribution of 3D content.
\end{itemize}

\section{Related Works}

\textbf{Text-to-3D: }Recent advancements in 3D generative models have been remarkable. Some methods~\cite{sanghi2022clip, liu2024uni3d,jain2022zero,mohammad2022clip,xu2023dream3d,liu2023unidream,zhang2024taming} leverage CLIP~\cite{radford2021learning} for 3D content generation. Subsequently, DreamFusion~\cite{poole2022dreamfusion} used cleverly designed loss functions to distill the powerful prior knowledge of 2D space diffusion models into 3D content. Prolificdreamer~\cite{wang2024prolificdreamer} proposed VSD and used LoRA~\cite{hu2021lora} to fine-tune pretrained diffusion models to fit the distribution between the input text and the 3D expression, thus improving the quality of the generated results. Magic3D~\cite{lin2023magic3d} applied a two-stage optimization strategy, first generating implicit representation and then refining the object's surface to generate a mesh. 

With the continuous development of high-quality 3D training datasets and computing resources, MVDream~\cite{shi2023mvdream} utilized the text-3D content pairs of the Objaverse~\cite{deitke2024objaverse,deitke2023objaverse} dataset to train a text-to-image diffusion model with multi-view consistency successfully. By integrating camera view information combined with score distillation sampling, MVDream~\cite{shi2023mvdream} can successfully generate 3D objects with multi-view consistency. DreamCraft3D~\cite{sun2023dreamcraft3d} and VP3D~\cite{chen2024vp3d} utilize multi-layer optimization to achieve higher visual fidelity and more detailed textures. GaussianDreamer~\cite{yi2024gaussiandreamer} quickly generates Gaussian from the text by bridging 2D and 3D diffusion models. Instant3D~\cite{li2023instant3d} trains a diffusion model capable of generating multi-view consistent content along with an improved transformer model. It uses the diffusion model to generate sparse views, then employs the transformer model to rapidly generate 3D objects based on these multiple views.

However, due to the limited semantic information in the text, it is often difficult to obtain the expected results in the generated outputs if the input text contains multiple objects.

\textbf{Image-to-3D: }Realfusion~\cite{melas2023realfusion}, Neurallift-360~\cite{xu2023neurallift}, and NeRDi~\cite{deng2023nerdi} use 2D diffusion models to generate images for 3D content. Anything3D~\cite{shen2023anything} integrates visual language models and object segmentation models to generate 3D textures and geometries from input images. Make-it-3D~\cite{tang2023make} employs a two-stage approach that combines textured point clouds with diffusion priors. Zero-1-to-3~\cite{liu2023zero} gains the ability to generate consistent multi-view images by fine-tuning a diffusion model. Magic123~\cite{qian2023magic123} combines 2D priors with 3D priors for generating 3D content. One-2-3-45~\cite{liu2024one} and One-2-3-45++~\cite{liu2023one} use a single image to quickly generate a 360-degree 3D mesh. SyncDreamer~\cite{liu2023syncdreamer} proposes a synchronous multi-view diffusion model to generate sparse views for 3D reconstruction, while Wonder3D~\cite{long2023wonder3d} applies a diffusion model to generate a consistent normal image and uses it for 3D generation. Splatter Image~\cite{szymanowicz2023splatter} uses a single image to generate Gaussian Splatting quickly. LRM~\cite{Hong_2023} and DMV3D~\cite{xu2023dmv3d} train with large datasets and use feedforward methods to reconstruct 3D quickly. However, these methods lack position control for generating multiple objects simultaneously. We address this issue by using bounding boxes.

\textbf{3D Controllability in Generation: }InstructNeRF2NeRF~\cite{haque2023instruct} utilizes an Image Conditional Diffusion Model (InstructPix2Pix~\cite{brooks2023instructpix2pix}) to iteratively edit input images while optimizing the underlying scene, thereby generating optimized 3D scenes that comply with editing instructions. Latent-NeRF~\cite{metzer2023latent} brings NeRF~\cite{mildenhall2021nerf} into latent space and uses sketch shapes to control content generation. Control4D~\cite{shao2023control4d} combines GAN~\cite{goodfellow2020generative} with ControlNet~\cite{zhang2023adding} to achieve control over portraits. SketchDream~\cite{liu2024sketchdream}, Sketch3D~\cite{zheng2024sketch3d} and  Control3D~\cite{chen2023control3d} achieve control through sketches. DreamControl~\cite{huang2024dreamcontrol} uses both ControlNet~\cite{zhang2023adding} and Conditional Lora to refine rough NeRF outputs. LucidDreaming~\cite{wang2023luciddreaming} controls voxels through clipped ray sampling and object-centric density bias initialization. DreamBooth3D~\cite{raj2023dreambooth3d} applies Dreambooth~\cite{ruiz2023dreambooth} fine-tuning to modify appearance, while IPDreamer~\cite{zeng2023ipdreamer} employs image prompts to provide specific appearance information for generated 3D objects. By contrast, our method achieves reasonable 3D multi-object generation control using simple 2D bounding boxes instead of these complex priors. In addition, compared to Build-A-Scene~\cite{eldesokey2024buildasceneinteractive3dlayout} that generates 2D images using text 3D layout, we focus on simultaneously generate multiple 3D objects.

\section{Method}

\begin{figure}[tbp]
\centering
\includegraphics[width=1.0\textwidth]{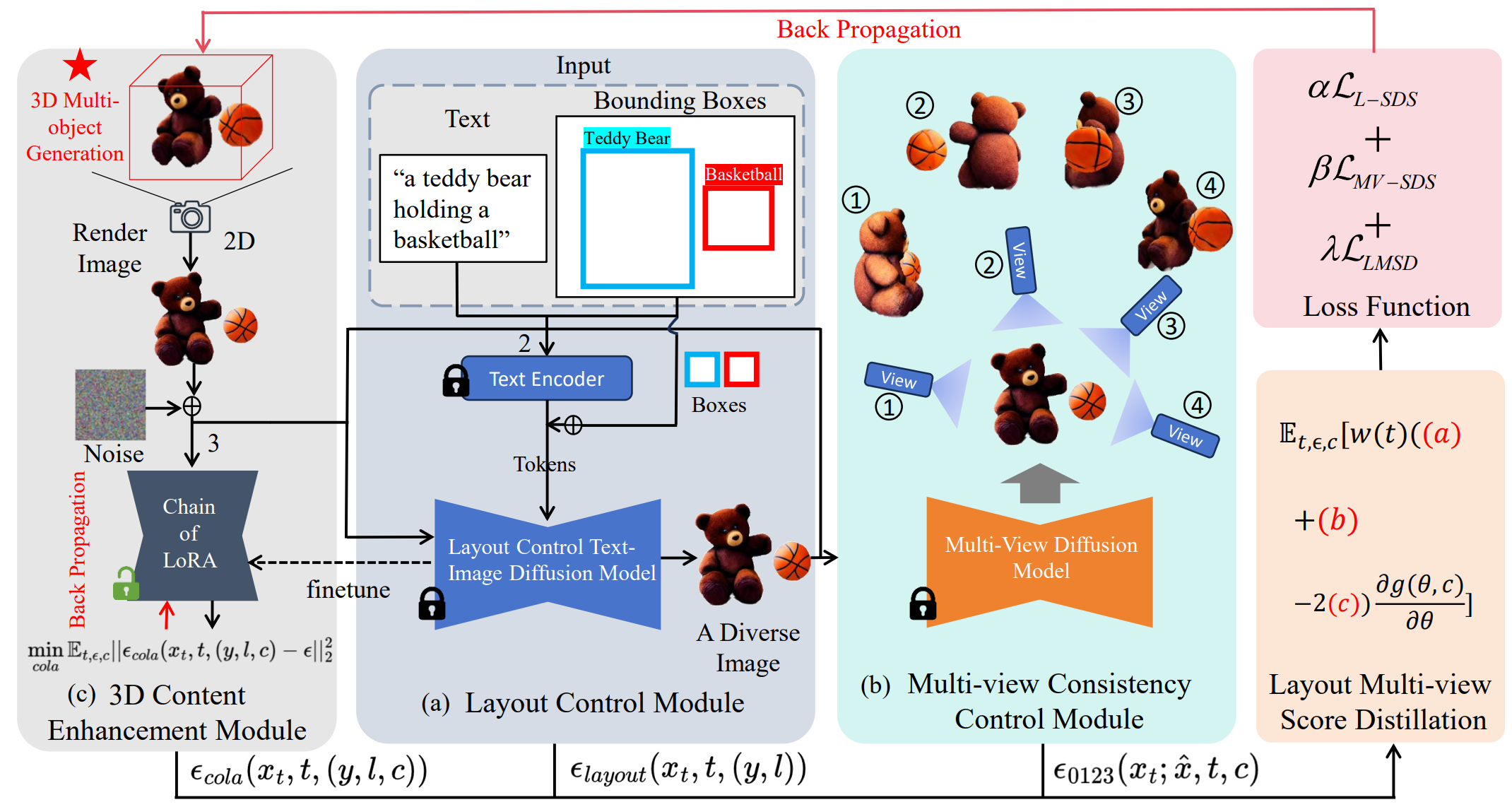}
\caption{An overview of the proposed COMOGen framework for controllable text-to-3D generation.}
\label{fig:2}
\end{figure}

We propose COMOGen for generating controllable 3D content from text and bounding boxes, as illustrated in Fig. \ref{fig:2}. 
COMOGen primarily consists of Layout Control Module (a), Multi-view Consistency Control Module (b), and 3D Content Enhancement Module (c). Firstly, COMOGen initializes 3D content and generates a 2D image based on the input text and bounding boxes, which serves as the 2D prior for the multi-view control module. In the Layout Control Module (a), it takes text and bounding boxes as guidance, using our designed Layout-SDS to distill layout prior knowledge from the Layout Control pretrained model. At the same time, the Multi-view Consistency Control Module (b) uses the 2D prior as guidance and leverages Multi-view SDS to distill pre-trained knowledge. Meanwhile, the 3D Content Enhancement Module (c) takes noisy images rendered from the 3D content to be optimized as input, using COLA loss to fine-tune the Layout Control diffusion model to fit the distribution of the generated 3D content. Finally, we jointly optimize multiple 3D objects by designing LMSD and computing multiple loss values.

\subsection{Preliminary}

\textbf{Latent Diffusion Model}: The diffusion-based approach is indeed one of the most effective model families for the text-to-image task, with the Latent Diffusion Model (LDM)~\cite{rombach2022high} and its subsequent Stable Diffusion being among the most powerful publicly available models in the research community. To mitigate the computational burden associated with training conventional diffusion models, LDM~\cite{rombach2022high} adopts a two-stage approach. In the initial stage, training involves a bidirectional mapping network tasked with transforming an image \( x \) into a latent representation, followed by reconstructing the image from this latent representation. Subsequently, in the second stage, a denoising model is trained within the latent space. Starting from \( x_T \), the model progressively generates smaller noise samples \( x_{T-1}, x_{T-2}, \dots, x_0 \), conditioned on a prompt \( y \) at each time step \( t \). At each step, LDM\cite{rombach2022high} employs a UNet model parameterized by \( \theta \), denoted as \( \epsilon_\theta \), to address the denoising challenge of the latent representation for the image \( x \). The optimization goal of LDM is:  
\begin{align}
\mathop {\min }\limits_\theta\
    \mathcal{L}_{\text{LDM}} = \mathbb{E}_{x,t,y,\epsilon\sim\mathcal{N}(0,\mathbf{I})}[w(t)||\epsilon-\epsilon_\theta(x_{t},t,y)||_2^2],
\end{align}
where \( t \) is uniformly sampled from the time steps \( \{1, \dots, T\} \), where \( x_t \) signifies the latent representation of the image targeted for denoising at time step \( t \). \( \epsilon \) denotes the noise introduced during the forward process, and \( \epsilon_\theta(\cdot, t, y) \) refers to the noise predicted by the UNet denoising network at time step \( t \), conditioned on the prompt \( y \).

\textbf{Score Distillation Sample}: Score distillation sample (SDS) is a method proposed by DreamFusion~\cite{poole2022dreamfusion} for distilling prior knowledge from 2D diffusion models. The core idea is to match the rendered image with the distribution modeled by the diffusion model and establish a connection between 3D representation and pretrained models using noise. Specifically, for a given text description, SDS first initializes a NeRF \(g(\theta)\) parameterized by \(\theta\). Through voxel rendering, it obtains a rendered image \(x=g(\theta ,c)\) at a given viewpoint \(c\). SDS adds noise \(\epsilon\) to this image and uses a pretrained UNet from the diffusion model to denoise it, observing the difference between these to judge the quality of the generated 3D content. Specifically, SDS calculates gradients for loss: 
\begin{align}
    &{\nabla _\theta }{\mathcal{L}_{SDS}}(\theta ) \approx {\mathbb{E}_{t,\epsilon,c}}[w(t)({\epsilon_{pretrain}}({x_t},t,y) - \epsilon)\frac{{\partial g(\theta ,c)}}{{\partial \theta }}].
\end{align}

\subsection{Layout Control Module}
In the Layout Control Module, we proposed Layout-SDS and applied GLIGEN~\cite{li2023gligen} to distill layout prior knowledge. Meanwhile, GLIGEN~\cite{li2023gligen} is utilized to generate images during the initialization phase, which serve as the prior for the multi-view control module. This approach enables both local and global control over the generated 3D content.

\textbf{Layout-SDS: }
In Layout-SDS, we expand the input of the original SDS to an instruction \((y, l)\), explored multimodal input. Here, \(y\) represents the caption, processed similarly to traditional methods. \(l\) denotes semantic information about grounded entities, such as bounding boxes. The position information of the bounding boxes is represented using coordinates, with each bounding box defined by two coordinates representing the top-left and bottom-right corners.
Next, the same pretrained text encoder is used to extract text features, and a MLP is employed to generate a grounding token \(h_l\) to represent \(l\). To achieve layout control, a gated self-attention layer is added between the self-attention and cross-attention layers within the UNet. Next, through voxel rendering, we obtain a rendered image \(x=g(\theta ,c)\) at a given viewpoint \(c\) from the 3D content \(g(\theta)\) initialized with parameter \(\theta\). Layout-SDS adds noise \(\epsilon\) to this image and uses \(\epsilon_{layout}\) as our denoiser to denoise it within the latent space, and the gradient of a Layout-SDS loss is calculated:
\begin{align}
    &{\nabla _\theta }{\mathcal{L}_{L-SDS}}(\theta ) \approx {\mathbb{E}_{t,\epsilon,c}}[w(t)({\epsilon_{layout}}({x_t},t,(y,l)) - \epsilon)\frac{{\partial g(\theta ,c)}}{{\partial \theta }}].
\end{align}

\subsection{Multi-view Control Module}
The above Layout-SDS can only obtain text and corresponding layout information from pretrained models, and we still face the Janus problem. To address this limitation, we introduce a Multi-view SDS that distills prior knowledge from the multi-view diffusion model, Stable Zero123~\cite{stale-zero123}, through using 2D prior image input from the layout control module as guidance, further enhancing the consistency of the generated 3D views and optimizing the 3D content, promoted Stable Zero123~\cite{stale-zero123} to multiple objects.

\textbf{Multi-view SDS: } 
Similarly, for MV-SDS, we first need to render an image \(x=g(\theta ,c)\) from the 3D representation \(g(\theta)\), in order to utilize perspective information \(c\), we apply the input image \(\hat x\) from the layout control module as a prior, using a denoiser \(\epsilon_{0123}\) from Stable Zero123~\cite{stale-zero123}, through denoise the noise image after adding noise to \(x\), using the removed noise \({\epsilon_{0123}}({x_t};\hat x,t,c)\) minus the added noise \(\epsilon\) as signals, we align the rendered view with the reference image in perspective \(c\) :
\begin{align}
    &{\nabla _\theta }{\mathcal{L}_{MV-SDS}}(\theta ) \approx {\mathbb{E}_{t,\epsilon,c}}[w(t)({\epsilon_{0123}}({x_t};\hat x,t,c)) - \epsilon)\frac{{\partial g(\theta ,c)}}{{\partial \theta }}].
\end{align}

\begin{figure}[t]
\centering
\includegraphics[width=1\textwidth]{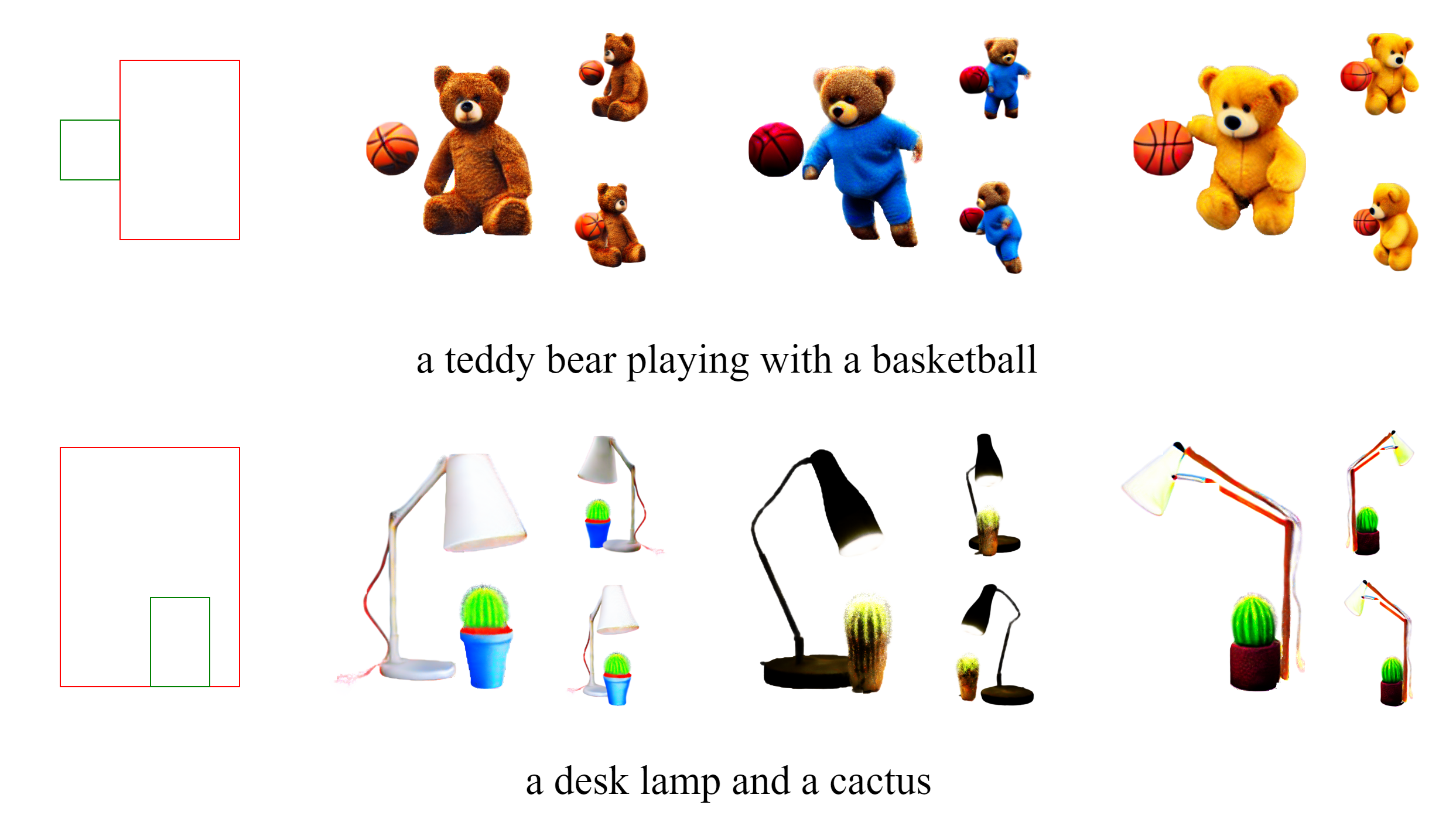}
\caption{Diversity generation samples of COMOGen.}
\label{fig:3}
\end{figure}
\subsection{3D Content Enhancement Module and Layout Multi-view Score Distillation }
The above Layout-SDS and Multi-view SDS face the same problems of SDS, such as over-saturation, over-smoothing, and low diversity. To further enhance the generation quality, we introduce 3D Content Enhancement Module to generate realistic and diverse 3D samples. Furthermore, in order to unify the prior knowledge of layout and multi-view, we introduce Layout Multi-view Score Distillation.

\textbf{3D Content Enhancement Module: } 
The 3D content enhancement module primarily comprises a fine-tuned diffusion model. Analogous to Variational Score Distillation~\cite{wang2024prolificdreamer}, our optimization objective aims to minimize the disparity between the pretrained distribution in the diffusion model corresponding to the input and the distribution of the 3D representation being optimized. 

To refine the distribution of 3D implicit representations, 
we specifically employ COLA (Chain of LoRA)~\cite{xia2024chain}, an iterative optimization framework tailored for parameter-efficient fine-tuning of large models. At the start of COLA, we use LoRA fine-tuning to denoise noisy images using input text and bounding boxes as guidance, and train our fine-tuning model using Eq.\ref{eq5}. Especially, for every 1000 steps of optimization, we integrate the learned low rank matrix into the fine-tuning model parameters. By iteratively integrating the learned low rank matrices into the model parameters, we can enhance our generalization ability without increasing computational or memory costs, so as to better fit the distribution of 3D representations.
\begin{align}\label{eq5}
\mathop {\min }\limits_{cola} {\mathbb{E}_{t,{\epsilon},c}}||{{\epsilon}_{cola}}({x_t},t,(y,l,c) - {\epsilon}||_2^2.\
\end{align}

\textbf{Layout Multi-view Score Distillation: }To unify the knowledge distilled from the layout diffusion model and the multi-view diffusion model, we designed LMSD , our aim is: 
\begin{align}
{\min {D_{KL}}[{q_t}({x_t}|c)||({p_{layout}}({x_t}|(y,l)) + {p_{mv}}({x_t}|(\hat x,c)))],}
\end{align}
where we use COLA to fine-tune the layout diffusion model to represent the distribution in the 3D representation for fitting the 3D representation \({q_t}({x_t}|c)\) with the layout distribution \({p_{layout}}({x_t}|(y,l))\) of the diffusion model under the given text with bounding box and the multi-view distribution \({p_{mv}}({x_t}|(\hat x,c))\),  not only enhances the consistency of layout and multi-view, but also improves the quality of the generated results. Furthermore, we use the removed noise to represent these distributions:
\begin{align}
    {\nabla _\theta }{\mathcal{L}_{LMSD}}(\theta ) \approx {\mathbb{E}_{t,\epsilon,c}}[w(t)({\epsilon_{layout}}&({x_t},t,(y,l))+{\epsilon_{0123}}({x_t};\hat x,t,c))\nonumber\\
    - {2\epsilon_{cola}}(x_t,t,&(y,l,c)))\frac{{\partial g(\theta ,c)}}{{\partial \theta }}].
\end{align}

\textbf{In General: }
COMOGen utilizes a layout control module and a multi-view control module to acquire layout prior knowledge and multi-view consistent prior knowledge through \( \mathcal{L}_{L-SDS} \) and \( \mathcal{L}_{MV-SDS} \) respectively. Finally, \( \mathcal{L}_{LMSD} \) is employed to integrate the two prior knowledge sources, and \( \alpha \mathcal{L}_{L-SDS} + \beta \mathcal{L}_{MV-SDS} + \lambda \mathcal{L}_{LMSD} \) is utilized to jointly optimize the 3D representation. Here, \( \alpha \), \( \beta \), and \( \lambda \) represent the trade-off parameters.

\section{Experiments}
In this section, we demonstrate the experimental results of COMOGen in generating controllable 3D content from text using bounding boxes. In Sec 4.1, we present the experimental setup. In Sec 4.2, we provide comparative experimental results with baselines in visual comparison results, CLIP score, user study and T3Bench~\cite{he2023t}. In Sec 4.3, we conduct ablation experiments to validate the effectiveness of our proposed modules.

\begin{figure}[t]
\centering
\includegraphics[width=1\textwidth]{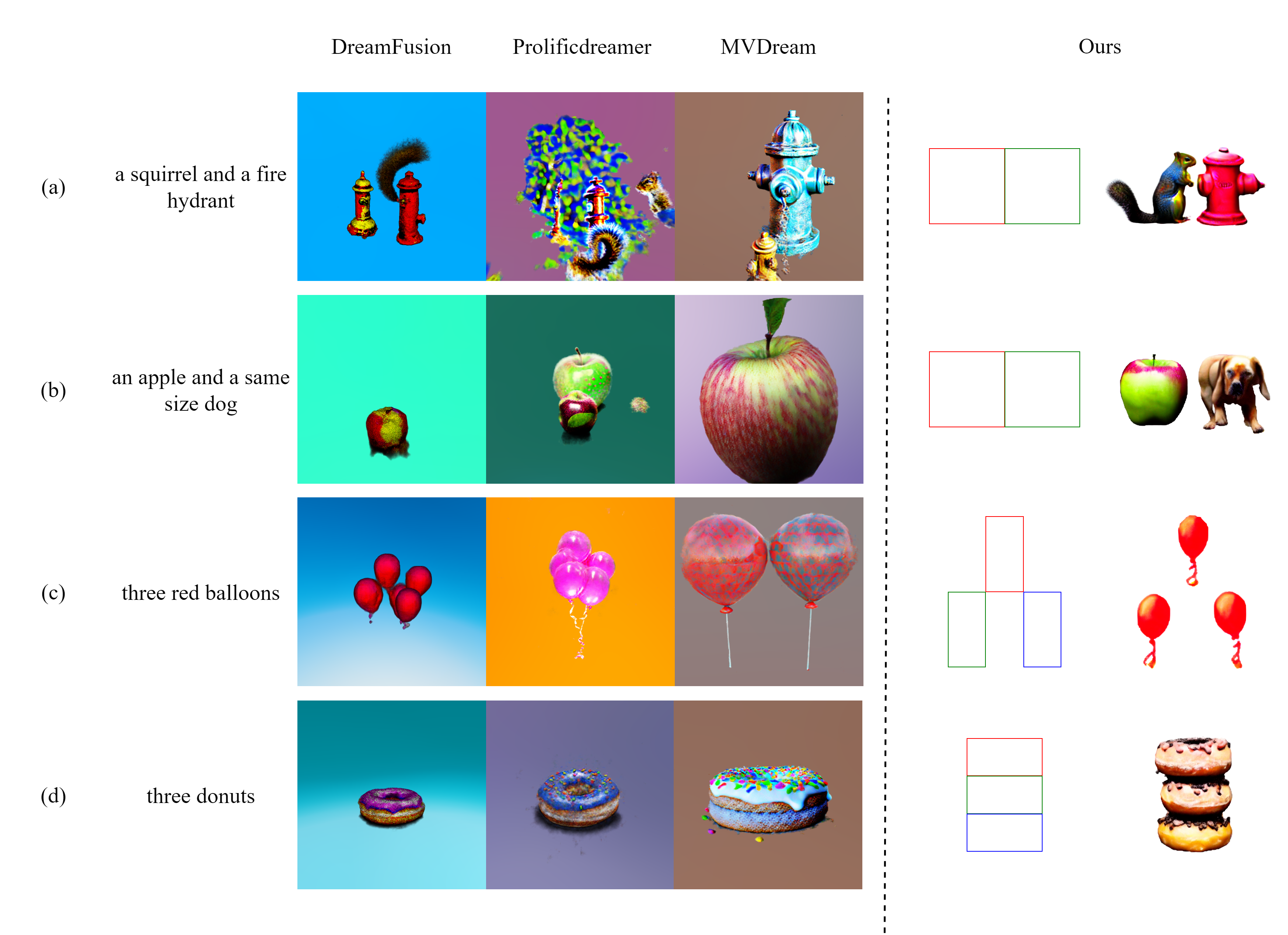}
\caption{Comparison with baselines. COMOGen can generate more reasonable 3D content.}
\label{fig:4}
\end{figure}

\subsection{Experiment setup}

\textbf{Baseline: }Since the proposed COMOGen is an optimization-based 3D content generation method, we chose several representative classic text-to-3D methods: DreamFusion~\cite{poole2022dreamfusion}, ProlificDreamer~\cite{wang2024prolificdreamer} and MVDream~\cite{shi2023mvdream}, as our baselines for comparative experiments.

\textbf{Implementation: }To ensure fairness, we implemented our baselines using the code from ThreeStudio~\cite{threestudio2023}, all experiments can be conducted on a single 4090 GPU. The rendering resolution for COMOGen and the baselines was limited to 128x128, with the training iterations set to 10,000 steps. For every method, we used 20 different text prompts to generate 60 3D contents. Each text prompt included multiple objects and various potential positional relationships. For COMOGen, we generated 3D contents with three different positional relationships for each text prompt, while for the baselines, we generated three 3D contents for the same text prompt.

\subsection{Comparative Experiments }
\textbf{Qualitative comparisions: }We show the experimental results of COMOGen in Fig. \ref{fig:1}. By distilling and integrating layout knowledge and multi-view consistency knowledge, the 3D objects generated by COMOGen can still generate reasonable 3D content when faced with text containing combinatorial concepts. Furthermore, through the application of the layout control module, the multi-view module and LMSD, COMOGen can also control the position of the generated object through the bounding box to make it as consistent as possible with our input expectations. We show the comparison results with the baseline in Fig. \ref{fig:4}. It can be easily found that the 3D content generated by these methods is mixed, and the objects are fused. While profit from three modules and LMSD, the generated results of COMOGen are more in line with the semantics of the input and do not contain Janus problems, which has significant advantages. At the same time, benefit from the 3D enhancement module and GLIGEN, the 3D content generated by COMOGen is rich in diversity, as shown in Fig. \ref{fig:3}.

\begin{table}[htbp]
\caption{Quantitative comparison of ours and baseline on CLIP score and user study.}
\label{tab:t1}
\centering
\begin{tabular}{ccc}
\toprule
Methods&CLIP Score\((\uparrow)\)&User study\((\uparrow)\)\\
\midrule
DreamFusion~\cite{poole2022dreamfusion}&30.08&3.59 \\
Prolificdreamer~\cite{wang2024prolificdreamer}&30.21&11.94 \\
MVDream~\cite{shi2023mvdream}&30.63&20.29 \\
COMOGen(Ours) & \bf 31.08& \bf64.17 \\
\bottomrule
\end{tabular}
\end{table}

\textbf{Quantitative evaluations: }
At first, we follow the evaluation method of Instant3D \cite{li2023instant3d} and use the CLIP score to assess the generated 3D content. To minimize the positive impact of the Janus problem on the CLIP score, we render views of each 3D content from three different perspectives (front, left-front, and right-front). We use the CLIP ViT-B/32~\cite{radford2021learning} to extract text and image features and calculate the CLIP score by averaging the similarity between each view and the input text prompt. The results in Table \ref{tab:t1} show that COMOGen has an advantage in CLIP score, indicating that it statistically outperforms baseline methods in text-to-multi-object generation and multi-object layout.
However, owing to the fact that Clip score simply renders 2D images from a specified perspective, it is inevitable that there will be some occlusion between objects in the 3D generated content of multiple objects, resulting in a decrease in clip score, a more reasonable quantitative metrics is still needed for multi-object generation.

To overcome the above issue and evaluate the text-to-multi-object more effectively, inspired by the quantitative experiments of VP3D~\cite{chen2024vp3d} and GaussianDreamer~\cite{yi2024gaussiandreamer}, we tested our method on T3Bench~\cite{he2023t} to make the experimental results more aligned with human objective evaluation.
The experimental results for multi-object text prompts are shown in Table \ref{tab:t2}. The experimental results on multi object text prompts are shown in Table 2. Compared with other methods, our method has achieved advantages in both quality and alignment, especially COMOGen has achieved a significant lead in alignment assessment.

In addition, we also conducted a user study experiment. We randomly selected 10 text prompts and generated 40 pieces of 3D content. The study used a multiple-choice format, where each question provided only the text prompt and the corresponding generated results from the four methods. 
We recruited 103 participants and instructed them to choose the content they felt best matched the prompt by considering critique text matching, texture quality of the 3D content, and positional relationships. As shown in Table \ref{tab:t1}, our proposed COMOGen received 64.17\% of the votes, making it the preferred choice. We achieved a significant 43.88\% accuracy gain compared to the second-ranked algorithm, MVDream~\cite{shi2023mvdream}. This outcome emphasizes the visual superiority of our approach.

\begin{table}[htbp]
\caption{Quantitative comparison on T3Bench~\cite{he2023t}.}
\label{tab:t2}
\centering
\begin{tabular}{cccc}
\toprule
Methods&Quality\((\uparrow)\)&Alignment\((\uparrow)\)&Average\((\uparrow)\)\\
\midrule
DreamFusion~\cite{poole2022dreamfusion}&17.3&14.8&16.1 \\
Prolificdreamer~\cite{wang2024prolificdreamer}&45.7&25.8&35.8 \\
MVDream~\cite{shi2023mvdream}&39.0&28.5&33.8 \\
VP3D~\cite{chen2024vp3d}&49.3&31.5&40.3 \\
GaussianDreamer~\cite{yi2024gaussiandreamer}&/&/&34.5 \\
COMOGen(Ours) & \bf 50.2& \bf43.8& \bf47.0 \\
\bottomrule
\end{tabular}
\end{table}

\subsection{Ablation Study}
In this section, we use quantitative analysis methods to evaluate the effect of L-SDS, MV-SDS, and LMSD. We use CLIP score, maintaining the initial settings of the experiment, and evaluate the experimental results under twenty text prompts with 60 generated 3D contents.

\textbf{Layout-SDS: }
In our method, Layout-SDS is used in the layout control module, aiming to distill prior knowledge from the layout control module, ensuring that the 3D content we finally generate meets our expectations. Additionally, the layout control module also provides 2D prior for the multi-view control module, which is used to maintain multifaceted consistency. The experimental results are shown in Table \ref{tab:t3}. In the absence of layout-SDS, due to the lack of layout prior knowledge, COMOGen is unable to accurately generate 3D content that conforms to complex text descriptions, resulting in a decrease in evaluation metrics.

\textbf{MV-SDS: } 
In COMOGen, MV-SDS is used to maintain multi-view consistency of 3D content, thus eliminating the Janus problem. In the ablation experiment, owing to the lack of MV-SDS constraints, the generated 3D content is difficult to maintain consistency across multiple views, resulting in a significant reduction in CLIP score, as shown in Table \ref{tab:t3}.

\textbf{LMSD: } 
Layout Multi-view Sample Distillation is designed to unify the prior knowledge distilled from the layout diffusion model and multi-view diffusion model, while using COLA fine-tuning to improve the quality of the generated 3D content. Table \ref{tab:t3} shows that LMSD can indeed improve the quality of 3D content while maintaining layout consistency and multi-faceted consistency.

\begin{table}[h]
\caption{Ablation study on CLIP score.}
\label{tab:t3}
\centering
\begin{tabular}{ccccc}
\toprule
Exp ID&L-SDS&MV-SDS&LMSD&CLIP Score\\
\midrule
(a)&\(\times\)&\(\checkmark\)&\(\checkmark\)&30.46 \\
(b)&\(\checkmark\)&\(\times\)&\(\checkmark\)&22.86 \\
(c)&\(\checkmark\)&\(\checkmark\)&\(\times\)&29.97 \\
(d)&\(\checkmark\)&\(\checkmark\)&\(\checkmark\)& \textbf{31.08} \\
\bottomrule
\end{tabular}
\end{table}

\section{Conclusion}
In this article, we propose COMOGen, a controllable 3D generation method capable of generating multiple objects from complex text and given bounding boxes. COMOGen is based on the layout diffusion model and multi-view diffusion model. The novel distillation method we propose extracts prior knowledge from pretrained models to analyze combined concepts. We have discussed and analyzed each module of COMOGen, and the experimental results emphasize the feasibility and superiority of our method. However, we have observed that widely used evaluation metrics do not effectively quantify multiple 3D objects. Although our visual comparison results significantly outperform the baseline, this superiority is challenging to demonstrate in quantitative experimental results. Additionally, our method has a limitation, i.e., the 2D bounding box we use cannot accurately describe the positional relationship on the z-axis, which will be addressed in future work.

\bibliographystyle{plain}
\bibliography{reference}

\end{document}